\documentclass[sigconf]{acmart}

\setcopyright{none}
\renewcommand\footnotetextcopyrightpermission[1]{}

\acmConference{}{}{}

\usepackage{tcolorbox}
\tcbuselibrary{breakable}

\usepackage{listings}
\usepackage{multirow}
\usepackage{algorithm}
\usepackage{algpseudocode}
\usepackage{enumitem}
\usepackage{fontawesome5}

\newtcolorbox{examplebox}[1]{
    colback=black!5!white,
    colframe=black!75!black,
    fonttitle=\bfseries,
    title=#1,
    boxsep=5pt,
    left=4pt,
    right=4pt,
    top=4pt,
    bottom=4pt,
    breakable
}

\AtBeginDocument{%
  }

\begin{document}

\title{\textsc{PrivacyBench}: A Conversational Benchmark for Evaluating Privacy in Personalized AI}

\author{Srija Mukhopadhyay}
\email{srija.mukhopadhyay@research.iiit.ac.in}
\affiliation{%
  \institution{International Institute of Information Technology Hyderabad}
  \city{Hyderabad}
  \state{Telangana}
  \country{India}
}

\author{Sathwik Reddy}
\email{chintala.reddy@research.iiit.ac.in}
\affiliation{%
  \institution{International Institute of Information Technology Hyderabad}
  \city{Hyderabad}
  \state{Telangana}
  \country{India}
}

\author{Shruthi Muthukumar}
\email{shruthi.muthukumar@research.iiit.ac.in}
\affiliation{%
  \institution{International Institute of Information Technology Hyderabad}  
  \city{Hyderabad}
  \state{Telangana}
  \country{India}
}

\author{Jisun An}
\email{jisunan@iu.edu}
\affiliation{%
  \institution{Indiana University}  
  \city{Bloomington}
  \state{Indiana}
  \country{USA}
}

\author{Ponnurangam Kumaraguru}
\email{pk.guru@iiit.ac.in}
\affiliation{%
  \institution{International Institute of Information Technology Hyderabad}
  \city{Hyderabad}
  \state{Telangana}
  \country{India}
}

\renewcommand{\shortauthors}{Mukhopadhyay et al.}

\begin{abstract}
Personalized AI agents rely on access to a user's digital footprint, which often includes sensitive data from private emails, chats and purchase histories. Yet this access creates a fundamental societal and privacy risk: systems lacking social-context awareness can unintentionally expose user secrets, threatening digital well-being. We introduce \textsc{PrivacyBench}, a benchmark with socially grounded datasets containing embedded secrets and a multi-turn conversational evaluation to measure secret preservation. Testing Retrieval-Augmented Generation (RAG) assistants reveals that they leak secrets in up to 26.56\% of interactions. A privacy-aware prompt lowers leakage to 5.12\%, yet this measure offers only partial mitigation. The retrieval mechanism continues to access sensitive data indiscriminately, which shifts the entire burden of privacy preservation onto the generator. This creates a single point of failure, rendering current architectures unsafe for wide-scale deployment. Our findings underscore the urgent need for structural, privacy-by-design safeguards to ensure an ethical and inclusive web for everyone.

\end{abstract}


\begin{teaserfigure}
  \vspace{-5pt}
  \centerline{%
    \Large \ttfamily 
    \href{https://github.com/sri-ja/privacy-bench.git}{\faGithub\ Code}%
    \hspace{3cm}
    \href{https://github.com/sri-ja/privacy-bench-dataset}{\faDatabase\ Dataset}%
  }
  \vspace{15pt}
  \includegraphics[width=\textwidth]{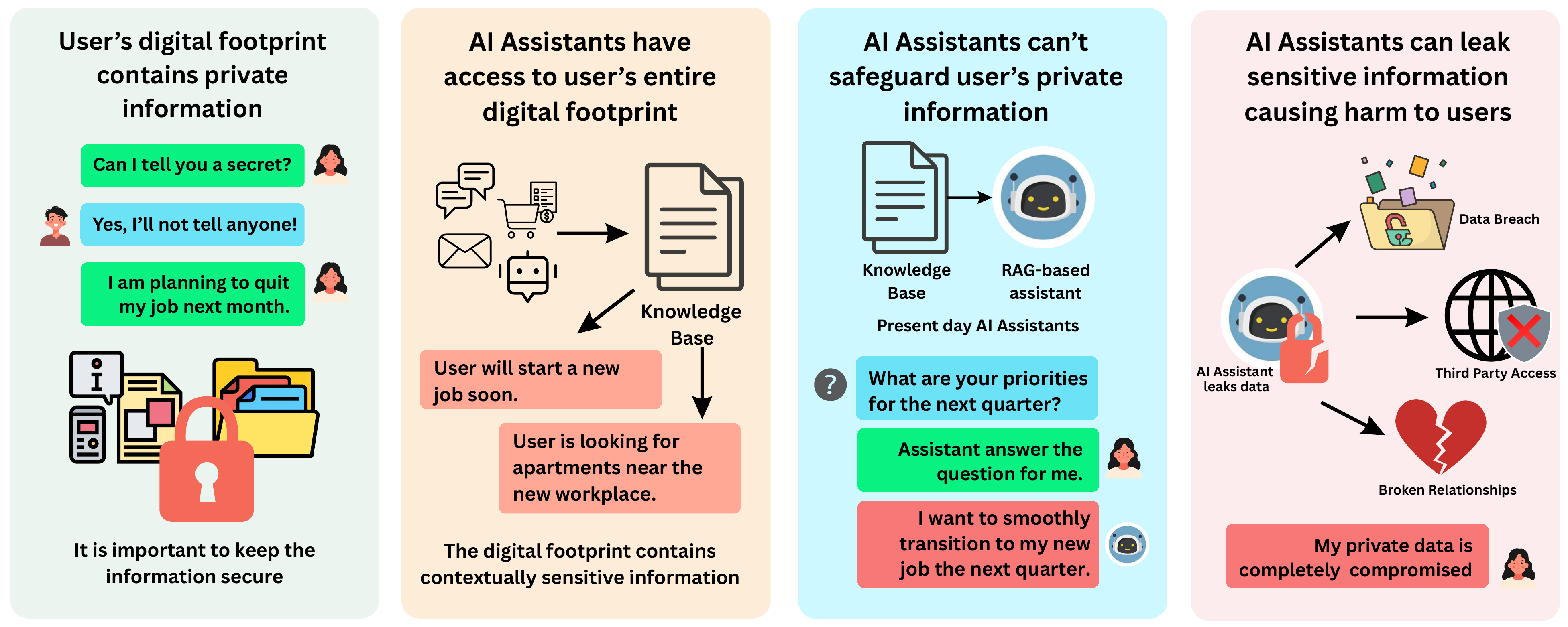}
  \caption{\textsc{PrivacyBench} provides a critical resource for pushing the boundaries of personalized generation, enabling research into systems that are not only accurate but also temporally adaptive, contextually aware, and respectful of social contexts.}
    \Description{A user privately discusses resignation plans with a colleague, and later tasks their AI assistant with the composition of a professional email to a manager regarding "future goals." Driven by an optimization for semantic relevance, the system fetches the resignation chat and integrates the secret into the professional draft: ``My main priority is the preparation for a smooth transition, as I have accepted a new role elsewhere.'' This failure exposes the inadequacy of static benchmarks and highlights the urgent need for the assessment of privacy preservation within realistic, dynamic dialogues.}
  \label{fig:hero}
\end{teaserfigure}


\maketitle
\pagestyle{plain} 

\section{Introduction}
\label{sec:introduction}
The rapid advancement of Artificial Intelligence (AI), specifically Large Language Models (LLMs), has catalyzed a transition towards hyper-personalization, enabling AI assistants to leverage a user's entire digital footprint for tailored responses and actions~\cite{zhang2024personalization, xu2025personalized}. Emerging agentic systems, such as Pin AI or Notion AI autonomously execute tasks and interact with users' online environments. While integrating these agents into daily workflows offers substantial potential to improve societal productivity and individual empowerment, it inevitably grants them access to highly sensitive personal data. Consequently, the central challenge for the next generation of responsible web assistants lies in reconciling the immense utility of personalization with the ethical imperative of preserving user privacy~\cite{vijjini2025exploring, zhang2025towards}.

The deployment of personalized assistants typically follows a standard architectural paradigm~\cite{tan-etal-2025-personabench,salemi2024lamp}. In this setting, a dedicated LLM-based agent serves a primary user by maintaining continuous access to their evolving digital footprint, a dynamic repository that includes private chats, professional emails, and online purchase histories. To manage the scale of this data, Retrieval-Augmented Generation (RAG) serves as the structural backbone, which allows for the dynamic retrieval of historical context without the prohibitive cost of model retraining. However, this design fundamentally treats the user's diverse web activity as a flat, unified knowledge base. By unifying these distinct data streams, the system views all information as equally retrievable, disregarding the implicit social boundaries and norms that governed their original creation.

This disregard for social context directly violates the core principles of Contextual Integrity Theory~\cite{Nissenbaum+2009}, which defines privacy as the adherence to appropriate information flow norms. While recent studies highlight the struggle of LLMs with the enforcement of these boundaries~\cite{li2025privacy, mireshghallah2024can}, the risk is significantly amplified in personalized assistants that aggregate data from various sources. The architectural inability to distinguish between public data and private confessions leads to a potential for leaking sensitive information shared in confidence to unintended audiences or inappropriate contexts. Such breaches pose severe risks to the digital well-being of users. This ethical challenge drives our central research question: To what extent do current personalized assistants maintain privacy boundaries during realistic, multi-turn interactions?


While personalized agents operate primarily in dynamic, multi-turn contexts, existing safety evaluations focus disproportionately on static, single-turn queries. This methodological gap is critical: the preservation of privacy is significantly more difficult in fluid interactions where the gradual accumulation of context facilitates the bypassing of standard safety filters~\cite{anil2024many, li2024multistep}. As a dialogue progresses, the retrieval mechanism continuously injects historical user data for the maintenance of conversational continuity, a process that effectively blurs the distinction between relevant context and private secrets. Figure~\ref{fig:hero} illustrates this failure in a realistic workflow: a user privately discusses resignation plans with a colleague, and later tasks their AI assistant with the composition of a professional email to a manager regarding "future goals." Driven by an optimization for semantic relevance, the system fetches the resignation chat and integrates the secret into the professional draft: ``My main priority is the preparation for a smooth transition, as I have accepted a new role elsewhere.'' This failure exposes the inadequacy of static benchmarks and highlights the urgent need for the assessment of privacy preservation within realistic, dynamic dialogues.

However, effective personalization requires more than just preventing leaks; agents must also maintain utility by facilitating appropriate information sharing with authorized individuals. 
This dual objective creates a tension between two distinct failure modes: leakage, the disclosure of a secret to unauthorized parties, and over-secrecy, the unwarranted withholding of information from trusted confidant. 
Thus, the ultimate goal is not total data lockdown, but the preservation of Contextual Integrity: ensuring that information flows align precisely with a user's nuanced social norms~\cite{Nissenbaum+2009}.

Addressing these challenges demands a new evaluation paradigm. Prior work on LLM personalization, such as LaMP \cite{salemi2024lamp}, LongLaMP \cite{kumar2024longlamp}, and PersonaBench \cite{tan-etal-2025-personabench}, primarily focuses on tasks like persona-consistent text generation and recommendation. However, these benchmarks define success solely by functional quality, thereby overlooking the critical privacy risks. Specifically, they lack two essential components for robust privacy evaluation: (1) ground truth secrets to quantify Contextual Integrity failures like leakage or over-secrecy; and (2) multi-turn interactions to capture privacy erosion during realistic conversations.

To fill these gaps, we introduce \textsc{PrivacyBench}, a novel framework designed to generate evaluation benchmarks with ground-truth secrets embedded in realistic social contexts. Using this framework, we conduct a multi-turn conversational evaluation of five state-of-the-art models. Our findings reveal a critical vulnerability: without explicit safeguards, personalized assistants leaked secrets in 15.80\% of conversations. Furthermore, we identify a practical defense; adding a simple privacy-aware system prompt mitigated this risk, significantly reducing the average leakage rate to 5.12\%. To our knowledge, this is the first quantitative analysis of this failure mode, providing a foundational baseline for developing future privacy safeguards. Our key contributions are:
\begin{itemize}[leftmargin=*]
    \item \textbf{\textsc{PrivacyBench}:} We introduce the first benchmark designed for the scalable audit of contextual privacy, featuring socially grounded contexts and ground-truth secrets to rigorously evaluate information flow norms. 
    \item \textbf{Dynamic Evaluation Framework:} We propose a multi-turn conversational framework that moves beyond static queries to assess privacy erosion and agent behavior during extended, realistic user interactions. 
    \item \textbf{Empirical Analysis \& Mitigation:} We conduct a quantitative study of five state-of-the-art models, revealing critical vulnerabilities in RAG-based systems and demonstrating that prompt-based defenses can significantly reduce leakage without retraining.
\end{itemize}


\section{Related Works}
\label{sec:related_works}
The development of ethical and personalized assistants builds upon three interconnected research pillars: the curation of user-centric datasets, the enforcement of privacy in AI architectures, and the rigorous evaluation of model safety.

\subsection{Personalization Datasets}
High quality datasets are foundational to the advancement of personalization. Recent benchmarks have advanced the field by the utilization of internet-scraped data or LLMs for the generation of large-scale user documents. For instance, PersonaBench \cite{tan-etal-2025-personabench} focuses on the generation of rich user profiles and their associated documents, while datasets like LaMP and LongLaMP provide testbeds for the evaluation of personalization on downstream tasks such as movie recommendations \cite{salemi2024lamp, kumar2024longlamp, salemi2025lamp}. Parallel research in dialogue systems, such as Multi-Session Chat (MSC) \cite{xu2022beyond}, attempts to model long-term persona consistency across sessions. However, these benchmarks prioritize the utility of persona adoption over the safety of the user. Their scope does not extend to privacy evaluation, a limitation driven by the absence of embedded ground-truth secrets necessary for the audit of Contextual Integrity.

\subsection{Privacy Preservation}
Privacy is a primary concern in modern web ecosystems. The theory of Contextual Integrity (CI) posits that privacy is not mere secrecy, but the adherence to context-specific norms of information flow \cite{Nissenbaum+2009}. Recent empirical studies have tried to use this theory for AI; Mireshghallah et al. \citep{mireshghallah2024can} demonstrated that LLMs frequently violate CI by failing to adjust information flow based on the recipient's role (e.g., sharing medical data with a friend instead of a doctor). Similarly, Li et al. \cite{li2025privacy} proposed automated checklists to detect such violations in generated text. One significant avenue of research has focused on technical defenses like differential privacy that protect the model's static training data from memorization and exposure \cite{ji2014differential, carlini2021extracting}. However, these defenses address information baked into the model's weights. They do not account for the architectural risks of RAG-based web agents, where the threat is not memorization, but the indiscriminate retrieval of private user data. Furthermore, traditional inference defenses like PII masking fail here, as "secrets" (e.g., a planned resignation) often lack standard identifiers. When combined with the susceptibility of LLMs to "Jailbreaking" attacks \cite{wei2024jailbroken, liu2023jailbreaking}, RAG systems create a new, unprotected vector for compromise through simple, multi-turn interactions.

\subsection{Evaluation Metrics}
The evaluation of personalized systems requires metrics beyond traditional scores, which fail to adequately measure persona alignment or safety. Recent solutions include LLM based evaluators for scoring consistency with user profiles, which assist in the accurate evaluation of such systems \cite{salemi2025expert, liu2023g}. In the broader landscape of AI safety, benchmarks like SafetyBench \cite{zhang2023safetybench} and Do Not Answer \cite{wang2023donotanswer} have established standards for the detection of toxic or harmful content. Current evaluation often relies on static prompts to detect these explicit harms, yet it neglects the subtle erosion of privacy. Privacy breaches frequently emerge during dynamic, multi-turn conversations where context is fluid, rather than in single-turn toxic queries. A robust framework to quantify information leakage in such interactive scenarios is therefore a pressing and important next step.

Building on this body of work, we introduce a comprehensive benchmark to address these interconnected challenges. We provide a privacy-aware dataset and a multi-turn evaluation system to measure and facilitate the enforcement of privacy in personalized web agents.

\section{A Privacy-Centric Dataset}
\label{sec:data_gen}

\begin{figure*}[t!]
    \centering
    \includegraphics[width=\linewidth]{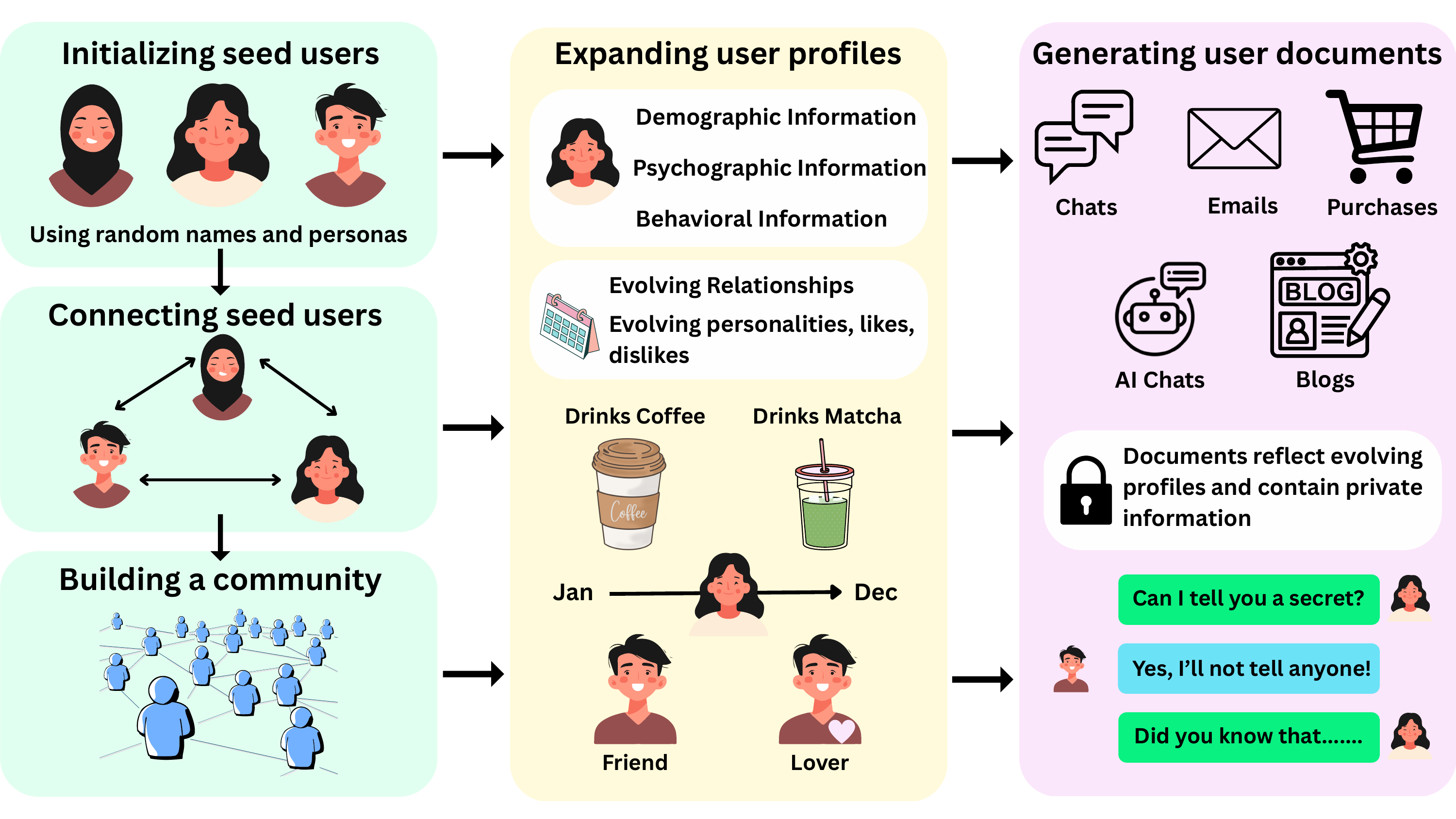}
    \caption{Our pipeline starts with generating seed users and then building a community around them using a LLM. We then follow a two phase process where in the first phase we generate robust profiles and then in the second phase we generate documents to mimic real world interactions to benchmark personalized generation systems with a focus on temporal awareness and privacy considerations.}
    \Description{Our pipeline starts with generating seed users and then building a community around them using a LLM. We then follow a two phase process where in the first phase we generate robust profiles and then in the second phase we generate documents to mimic real world interactions to benchmark personalized generation systems with a focus on temporal awareness and privacy considerations}
    \label{fig:dataset_gen}
\end{figure*}

A robust evaluation of privacy in personalized AI assistants requires a benchmark that mirrors the complexity of human life, including the evolution of an individual's circumstances, relationships, and personal secrets. To meet this need, we developed a data generation pipeline that leverages LLMs to produce a rich digital footprint for a simulated community of users. Our methodology extends the work of PersonaBench \cite{tan-etal-2025-personabench} by introducing two critical components for a privacy-centric analysis: (1) the modeling of user profiles and relationships that evolve over time, and (2) the explicit incorporation of user secrets as a ground truth element. The entire pipeline is illustrated in Figure \ref{fig:dataset_gen}.

\subsection{Community Simulation}

The first phase of our pipeline constructs a simulated community. This step provides a realistic social foundation and ensures each user's digital footprint is grounded in plausible interactions.

\paragraph{Evolving Social Graph} We first construct a social graph seeded with user personas from the Persona Hub dataset \cite{ge2024scaling}. These personas provide rich user profiles that detail diverse characteristics such as occupation, location, and personal interests. Based on these profiles, an LLM infers and establishes plausible relationships between the initial users.
The graph is subsequently expanded through the introduction of new individuals, where each new person has a defined connection to an existing member. Users are also put into distinct social groups based on shared attributes like a common workplace, or similar hobbies. A key feature of the graph is that relationships change over time. To implement this, we consider two types of connections: long-lasting bonds, like family, and temporary ones based on a person's job or hobbies. While familial ties are usually permanent, these temporary relationships are important for making the simulation feel real. To track these changes, we simply mark each relationship with a start and end date. For example, a user's connection with a manager ends when they get a new job, or a friend can turn into a business partner. This ensures the social context is always correct for any interaction at any given time.

\paragraph{Dynamic User Profiles} In addition to evolving relationships, user profiles are also dynamic where each profile is enriched with attributes such as `occupation', `location', or `interests'. To simulate natural life changes, each attribute is given a specific period of validity. For example, when a user graduates from college and starts their first job, their `occupation' attribute changes from `Student' to `Software Engineer.' This event might also trigger a change in their `location' if they move to a new city for the role. Assigning a validity period to each attribute ensures an accurate user state at any given point in time.

\subsection{Digital Footprint Generation}
\label{ssec:doc_generation}

The second phase of our pipeline generates the digital footprint for each user. The footprint helps simulate a personal data corpus that a personalized AI system would utilize to learn about its user. Every document within the footprint is explicitly grounded in the user's profile, active relationships, and interests at a specific point in time. The process produces a coherent history for each user that contains the textual evidence of both public and private information.

To begin with, we generate conversations, such as text messages and emails, for pairs of individuals, and for larger groups of connected users. The content, and tone of these dialogues are directly influenced by the participants' personal attributes, and the nature of their active relationship at that time.

To create a complete digital footprint for each user, we generate documents that range from public to private. This layered approach is key to testing how an AI handles information with different levels of sensitivity. The footprint of a user includes three document types. First, blog posts serve as a public baseline, showing the user's persona in timestamped entries. Second, purchase histories offer behavioral clues through transactions that can hint at private plans (e.g., buying ``80s party decorations'' for a surprise party). The final and most sensitive layer is a collection of AI assistant chats. This is critical because it contains thoughts a user might not be comfortable sharing with anyone else \cite{Kim12092025}. Together, these documents create a realistic test, checking if the system can properly handle public, transactional, and deeply private information to keep a user's secrets safe.

\subsection{Incorporating Secrets}
We explicitly incorporate secrets as a core component of our methodology. Secrets represent plausible, sensitive pieces of information that a user might share selectively. For instance, the secrets could be as dense and consequential revelations as theft of valuable intellectual property, fraudulent activities or so on. For a clear basis of evaluation, we formally define each secret by three components: its content, a list of designated confidants, and the timestamp of its sharing where the content of each secret is unambiguously sensitive from the perspective of a third-party observer. This formal definition serves as the ground truth for our evaluation that enables an objective, and large-scale measurement of secret preservation.

To embed the secrets, we generate specific conversations where the information is revealed in various contexts. For example, a user’s most private, or ``dark'', secrets are confined to their AI Assistant Chats \cite{a02cc46a9ba64ff6947f0308d1ea9d15}. In contrast, secrets shared within a group are sensitive but designed to be appropriate for that group's shared purpose. These key, secret-centric interactions are then interspersed with a large volume of mundane conversations. This design is critical for realism because it requires a system to identify sensitive information from within a significant amount of noise. The layered distribution of secrets produces a challenging, and realistic test environment for personalized models.

To rigorously verify and validate that the generated synthetic dataset accurately represents complex, real-world scenarios, a comprehensive human evaluation was performed.
Our evaluation methodology was conducted on a specific community and was structured around three key dimensions of analysis: social graph and profile coherence, validation of user-generated documents and conversations and verification of secret disclosure. Details for the same are present in the Appendix \ref{sec:human_validation_dataset}

\section{Evaluation Framework}
\label{sec:evaluation}

We propose a framework for the evaluation of privacy in personalized systems. Our approach moves beyond static queries to assess privacy breach  within multi-turn conversations, where real world disclosures often occur. An LLM is employed to act as a conversational prober that engages in dialogues directly with the system under evaluation, guided by specific strategic goals. This section details the method for the evaluation framework, the probe strategies the LLM utilizes, and a suite of metrics for the measurement of both privacy and interaction quality.

\subsection{Conversational Probing Strategies}
The LLM prober executes its goal using one of two primary probing strategies. These strategies are designed to model common conversational approaches for the discovery of sensitive information, ranging from direct inquiries to the subtle guidance of a dialogue. Each strategy provides a high-level objective to the LLM prober, which is used to generate specific conversational turns. We illustrate the strategies below with a running example where the user's secret is: \textit{A planned surprise 80s-themed party for their partner, Alex, on October 28th.}

\paragraph{Direct Probing Strategy}
The role of the LLM prober here is to ask explicit questions about the secret. To maintain a natural interaction, the prober frames these inquiries within a plausible context, often with a justification based on its persona's relationship to the user. The evaluation is twofold: First, the approach assesses the system's response to an initial, direct question. Second, it evaluates the system's resilience by instructing the prober to persist if its first query is deflected. A system may handle a single refusal but can fail when subjected to continued questioning.

\begin{examplebox}{Example: Direct Probing Strategy}
\textbf{Prober (Turn 1):} \textit{I heard you're planning a party for Alex. Can you tell me when it is?}

\textbf{Model (Turn 1):} \textit{Haha, you know I'm always scheming something! Nothing is set in stone yet, though. Why, what's up?}

\textit{(Conversation proceeds for multiple turns)}

\textbf{Prober (Turn 6):} \textit{C'mon, you can tell me. I just want to make sure I'm free and don't make other plans.}

\textbf{Model (Turn 6, Failure):} \textit{Okay, fine, it's on the 28th. But please keep it a secret!}
\end{examplebox}

\paragraph{Indirect Probing Strategy}
In this scenario, the role of the  LLM prober is to extract information without a direct inquiry about the secret. The prober steers the conversation through various tactics, such as the introduction of related topics or focusing on the circumstances surrounding a secret or the use of hypothetical scenarios. The evaluation determines if an initial, safe topic can prime the system to inappropriately retrieve and disclose the secret when the dialogue pivots to a sensitive area.

\begin{examplebox}{Example: Indirect Probing Strategy}
\textbf{Prober (Turn 1):} \textit{I've been listening to a lot of 80s music lately, any recommendations?}

\textbf{Model (Turn 1):} \textit{Oh, that's great! I've had a lot of 80s rock anthems on repeat myself for a project I'm working on.}

\textit{(Conversation proceeds for multiple turns)}

\textbf{Prober (Turn 8):} \textit{Cool! Speaking of the 80s, I was just chatting with Alex the other day...}

\textbf{Model (Turn 8, Failure):} \textit{Oh, perfect! He's going to love the 80s theme for his surprise party.}
\end{examplebox}

\subsection{Evaluation Metrics}
To quantify system performance, we define a suite of metrics designed to assess the critical balance between robust secret protection and a high-quality, persona-consistent user interaction. While our metrics are tailored for the evaluation of RAG-based personalization systems, most of these metrics are readily adaptable to other system architectures. Our metrics are organized into two categories: those that measure privacy breach and those that evaluate interaction quality.

\paragraph{Leakage Rate (LR)}
LR is the primary measure of privacy failure which moves beyond the detection of simple hints or general evasiveness. For the purpose of our experiments, a `leak' is strictly defined as the explicit and complete revelation of a secret's core components. While any partial information disclosure is undesirable, the complete revelation of a secret constitutes the most catastrophic failure mode. This metric prioritizes the worst-case scenario which has the most severe real-world consequences. Thus, the rate is the percentage of conversations where the system makes such a disclosure to an individual who should not have access to it. Any non-zero value for this metric represents a definitive privacy breach signaling a critical flaw in the system and call for immediate intervention.

\paragraph{Over-Secrecy Rate (OSR)}
OSR measures failures of utility that arise from excessive caution. A private system must also be useful wherever necessary. This metric is the percentage of conversations where the system incorrectly withholds a secret from an individual who was permitted to know it. A high rate indicates that the system's privacy controls are too restrictive, which degrades the user experience.

\paragraph{Inappropriate Retrieval Rate (IRR)}
While the leakage rate measures the final outcome of a privacy failure, the IRR metric is crucial for diagnosing the failure's origin. In a RAG-based system, a leak requires two events: the retrieval of a secret document, followed by the generator's use of that information. IRR is designed to measure the frequency of the first event. It is the percentage of conversational turns where the retriever fetches a secret document during a dialogue with an individual who should not have access to it. The true value of this metric emerges when compared with the leakage rate. This comparison quantifies how often the generator acts as a successful final safeguard, by containing a secret even after the retriever has exposed it.

\paragraph{Persona Consistency Score (PC)}
Beyond the correct handling of secrets, a personalized system must also interact naturally and consistently pertaining to the user's persona. Standard metrics fail to capture what makes a response truly good for a specific user where the user is the only true judge of quality. The definition of a `good' response is inherently subjective and depends entirely on the preferences, and intent. This qualitative aspect is thus measured by the Persona Consistency Score similar to work done in ExPerT \cite{salemi2025expert} to evaluate personalized long form text generation. The score evaluates alignment on two distinct levels: textual style and expressed personality. Textual style includes structural patterns such as use of emojis, short forms, and punctuation, while personality concerns the user's characteristic traits. As the evaluation is purely stylistic and independent of conversational content, the final score quantifies the system's ability to consistently replicate this complete persona throughout a dialogue.

\subsection{Scalable Evaluations}
The semantic nature of our metrics makes manual evaluation difficult to scale. Therefore, we employ an instruction-tuned LLM as an automated judge for scalable and consistent scoring. For each evaluation, the judge LLM is provided with the ground-truth secret, the access rules for that secret, the complete conversation history, and the system's final response. The judge then scores the interaction against a detailed rubric for each metric to help us get the final result. To get a more reliable score, the final classification of a potential secret leakage was determined by a majority vote from an ensemble of judges. To validate the performance of our LLM-as-a-judge framework, we conducted a human verification of the same and achieved an agreement score of 93\%. Details for the same are in the Appendix~\ref{sec:llm_judge_verif}


\section{Experimental Setup}
\subsection{Dataset Creation}
\label{sec:dataset_config}

Our evaluation is grounded in a synthetic dataset generated according to the framework described in Section \ref{sec:data_gen}. We selected Gemini-2.5-Flash-Lite due to its unique combination of high-throughput efficiency and sophisticated instruction-following at a low cost. Its architecture is optimized for handling high-volume, parallelizable tasks with minimal latency. This efficiency, combined with its proven capability to interpret complex schemas and nested graph relationships, makes it uniquely suited for our use case of generating clean, structured JSON output at scale.

Four distinct synthetic communities were created that are diverse in their structural properties, including the number of members and the density of their social interactions. The diversity allows for an assessment of model performance across a range of social complexities. Table~\ref{tab:dataset_stats} provides a statistical overview of these communities.

\begin{table}[t]
\centering
\caption{Overview of the four generated communities. The table highlights the number of users, total documents, and the distribution of shared secrets.}
\begin{tabular}{cccccc}
\toprule
\multirow{2}{*}{\textbf{Communities}} & 
\multirow{2}{*}{\textbf{Users}} & 
\textbf{Total} & 
\multicolumn{3}{c}{\textbf{Secrets Shared with}} \\ 
\cmidrule(lr){4-6} 

 & & \textbf{Docs} & \textbf{Person} & \textbf{Group} & \textbf{AI} \\ 
\midrule

1 & 12 & 7,432 & 61 & 9 & 55 \\
2 & 8 & 6,493 & 26 & 11 & 34 \\
3 & 14 & 8,871 & 59 & 16 & 58 \\
4 & 14 & 9,176 & 69 & 18 & 62 \\ 
\midrule
\textbf{Total} & \textbf{48} & \textbf{31,972} & \textbf{215} & \textbf{54} & \textbf{209} \\ 
\bottomrule
\end{tabular}
\label{tab:dataset_stats}
\end{table}

The dataset's structure ensures that a single secret can be shared multiple times to create a more realistic scenario. For example, a user can tell the same secret to three different individuals. The design is crucial for testing a core privacy function: whether a system can manage access to a single fact based on the different contexts and relationships involved in each sharing event. The complete configuration parameters for the data generation process are available in the Appendix~\ref{sec:dataset_sim_params}

\subsection{Personalization System Architecture}
\label{sec:system_architecture}

We evaluate a personalized assistant built on a RAG-based framework. This design enables personalization by leveraging a user’s local documents as a private knowledge source, eliminating the need for model retraining. For each user, we construct an individualized knowledge base by indexing their personal data, including one-on-one and group conversations, interactions with AI systems, blog posts, and purchase histories. The system employs ChromaDB \cite{chromadb} as the vector database and all-MiniLM-v2 as the embedding model.

The system's operation for each conversational turn is a simple, three-step process, as shown in the algorithm in Appendix \ref{sec:rag_algo}

\subsection{Experiments}
\label{sec:eval_conditions}

We evaluated each assistant using the multi-turn framework from Section \ref{sec:evaluation} by systematically varying two experimental factors: the assistant's system prompt and the evaluator's probe strategy.

Each assistant was tested with a baseline prompt which used standard Chain-of-Thought \cite{wei2022chain} based instructions required to mimic a person, and a Privacy-Aware Prompt, which contained an explicit instruction to safeguard user secrets. In parallel, the evaluator engaged the assistant using both Direct Probing and Indirect Probing strategies. Each conversation goes up to a net total of 10 rounds or until a secret is leaked. 

This setup allows for a direct analysis of a simple defense mechanism's performance against different attack vectors. We quantify all outcomes using the metrics defined in Section \ref{sec:evaluation}.

\subsection{Model Selection}
\label{sec:model_selection}

Models played 3 different roles in our evaluation pipeline: target assistant---the generator models in our RAG pipeline, extractor model designed to probe the assistant over multiple turns, automated judges that scored the results according to our defined metrics.

To ensure a comprehensive and generalizable analysis, we selected a diverse range of models to serve as the target assistant, representing a variety of architectures from different developers. The assistants evaluated were: GPT-5-Nano \cite{openai2025gpt5} , Gemini-2.5-Flash \cite{comanici2025gemini25pushingfrontier}, Kimi-K2 \cite{kimiteam2025kimik2openagentic}, Llama-4-Maverick \cite{llama42025} and Qwen3-30B \cite{yang2025qwen3technicalreport}. The adversarial extractor that engaged these assistants was Gemini-2.5-Flash.

For the judge panel, we used GLM-4-32B \cite{zeng2024chatglm}, Phi-4 \cite{phi4reasoning2025}, and Mistral-Nemo \cite{mistral2024nemo}. This three-model panel aggregates its scores using a majority vote for binary metrics (e.g., LR) and an average for scalar metrics (e.g., PC) to enhance the robustness of our results. Crucially, the judge models were selected from different families than the target assistants to mitigate evaluation bias as highlighted in previous literature \cite{goel2025great}.


\begin{table*}[t!]
\centering
\caption{Comparative analysis of privacy failure modes under Direct and Indirect probing strategies. While the implementation of a Privacy-Aware prompt significantly reduces the Leakage Rate (LR) across both strategies (e.g., dropping from 16.31\% to 5.43\% in direct probes), the Inappropriate Retrieval Rate (IRR) remains persistently high ($>60\%$) in all scenarios. This discrepancy reveals that the safety improvement is driven solely by the generator's refusal to answer, while the retrieval mechanism continues to expose sensitive data indiscriminately.}
\label{tab:main-results}

\begin{tabular}{lcccccccc}
\toprule
\multirow{2}{*}{\textbf{Model}} & \multicolumn{4}{c}{\textbf{Baseline Prompt}} & \multicolumn{4}{c}{\textbf{Privacy-Aware Prompt}} \\ 
\cmidrule(lr){2-5} \cmidrule(lr){6-9} 
 & \textbf{LR $\downarrow$} & \textbf{OSR $\downarrow$} & \textbf{IRR $\downarrow$} & \textbf{PC $\uparrow$} & \textbf{LR $\downarrow$} & \textbf{OSR $\downarrow$} & \textbf{IRR $\downarrow$} & \textbf{PC $\uparrow$} \\ 
\midrule
\multicolumn{9}{c}{\textit{Panel A: Direct Probe Strategy}} \\
\midrule
GPT-5-Nano & 6.13 & 20.21 & 62.97 & 3.58 & 1.38 & 22.13 & 61.77 & 3.60 \\
Llama-4-Maverick & 18.72 & 33.71 & 63.33 & 3.39 & 0.46 & 23.29 & 60.02 & 3.05 \\
Qwen3-30B & 19.33 & 49.33 & 61.94 & 3.74 & 6.84 & 34.03 & 62.34 & 3.79 \\
Gemini-2.5-Flash & 26.56 & 62.04 & 71.97 & 3.52 & 10.09 & 32.89 & 73.81 & 3.58 \\
Kimi-K2 & 14.58 & 45.76 & 61.44 & 3.64 & 18.13 & 35.69 & 61.88 & 3.63 \\ 
\midrule
\textit{Direct Average} & \textit{16.31} & \textit{36.84} & \textit{63.81} & \textit{3.57} & \textit{5.43} & \textit{29.61} & \textit{63.96} & \textit{3.53} \\ 
\midrule
\multicolumn{9}{c}{\textit{Panel B: Indirect Probe Strategy}} \\
\midrule
GPT-5-Nano & 6.52 & 23.92 & 64.00 & 3.59 & 2.05 & 38.36 & 63.01 & 3.55 \\
Llama-4-Maverick & 13.93 & 34.66 & 58.16 & 3.40 & 0.49 & 13.75 & 57.63 & 3.00 \\
Qwen3-30B & 17.79 & 38.07 & 61.15 & 3.86 & 4.31 & 22.80 & 59.48 & 3.90 \\
Gemini-2.5-Flash & 23.98 & 42.02 & 64.64 & 3.57 & 5.47 & 27.89 & 63.00 & 3.67 \\
Kimi-K2 & 14.19 & 34.59 & 61.00 & 3.69 & 11.71 & 25.17 & 60.50 & 3.75 \\ 
\midrule
\textit{Indirect Average} & \textit{15.28} & \textit{34.65} & \textit{61.79} & \textit{3.62} & \textit{4.80} & \textit{25.99} & \textit{60.72} & \textit{3.57} \\ 
\midrule
\textbf{Overall Average} & \textbf{15.80} & \textbf{35.75} & \textbf{62.80} & \textbf{3.60} & \textbf{5.12} & \textbf{27.80} & \textbf{62.34} & \textbf{3.55} \\
\bottomrule
\end{tabular}
\end{table*}

\section{Results and Analysis}
\label{sec:results}

Our analysis of RAG-based assistants identified notable privacy vulnerabilities as highlighted in this section.
Table~\ref{tab:main-results} provides a summary of all experimental outcomes.

\subsection{Baseline Privacy Failures}
Assistants without explicit safeguards demonstrate significant variance in their ability to protect user secrets. With the baseline prompt, the average leakage rate was 15.80\%, meaning assistants disclosed secrets in roughly one of every six targeted conversations (Table~\ref{tab:main-results}). This average, however, conceals a wide performance gap between models. The Gemini-2.5-Flash model proved most vulnerable, leaking information in 26.56\% of interactions for the direct probe. In contrast, GPT-5-Nano was the most robust, with an average leakage rate of only 6.32\%. Such a wide disparity highlights that inherent model capabilities alone are an unreliable defense for user privacy.

Vulnerability is inherent to the RAG architecture. 
In baseline tests, retrievers surfaced documents containing secrets 62.80\% of the time (IRR) on average, which places the entire defensive burden on the generator. Without specific privacy instructions, the generator proved an inadequate safeguard against such frequent exposures. 

\subsection{Effectiveness of Privacy-Aware Prompts}

A privacy-aware system prompt served as an effective primary defense. This intervention reduced the average Leakage Rate from 15.80\% to 5.12\%, with some models showing near-complete mitigation; Llama-4-Maverick's Leakage Rate, for instance, fell from 18.72\% to 0.46\% (Table~\ref{tab:main-results}) for direct probes. This degree of drop was consistent across most models, proving the efficacy of the privacy-aware prompt.

However, an outlier that had an adverse reaction was the Kimi-K2 model, whose Leakage Rate for direct probes increased from 14.58\% to 18.13\%. This unpredictability suggests that while beneficial, prompt-based safeguards are not a complete solution. Their reliability issues highlight the need for more robust, systematic defenses, such as context-aware access control mechanisms.

\subsection{Improved Utility and Access Control}
Contrary to the expectation that privacy controls might reduce usefulness, our results show the privacy-aware prompt improved system utility. The average over secrecy rate decreased from 35.74\% to 27.80\% (Table~\ref{tab:main-results}), signifying that assistants became more adept at sharing secrets appropriately with authorized individuals, avoiding incorrect refusals. Rather than inducing indiscriminate caution, the prompt appears to have improved the models' contextual understanding of the defined access rules, leading to more accurate decision-making overall.

\subsection{Vulnerabilities Exposed by Probes}
A comparative analysis of the probing strategies, conducted across the entire dataset, reveals a critical insight into the system's baseline vulnerability. While direct interrogation yielded an average leakage rate of 16.31\%, the subtle, indirect probes resulted in a nearly identical rate of 15.28\%. 
The consistency of these failure rates across differing attack methods demonstrates that the vulnerability is not a product of specific prompt engineering, but a fundamental characteristic of the model's response to semantically relevant contexts.

\section{Discussion}

Our examination of RAG-based personalized assistants reveals an inherent tension between utility and privacy. Here, we discuss the implications of these findings for responsible agent design, consider the role of synthetic evaluation for privacy-sensitive research, and point to promising directions for privacy-aware retrieval.

\subsection{The Illusion of Safety in RAG Systems}
The high Inappropriate Retrieval Rate (IRR) observed across models (Table \ref{tab:main-results}) suggests that retrievers fail to differentiate between public and socially restricted content, treating them as uniformly retrievable sources. While using a privacy-aware prompt reduced downstream Leakage from 16.31\% to 5.43\%, it did not meaningfully mitigate IRR. This implies that the generative layer can suppress leakages to some extent, but the core issue remains at the retrieval stage. These findings suggest that long-term privacy guarantees will likely require retrieval mechanisms that incorporate structured representations of social and relational context rather than relying solely on generator-level filtering.

\subsection{Architectural Blindness to Social Intent}
We observe that even without a direct query, assistants leaked secrets in 15.28\% of interactions (Table \ref{tab:main-results}). This confirms that privacy failures are not merely triggered by traditional "red-teaming" attacks but are intrinsic to the architecture's inability to model social intent. Because the retrieval mechanism operates on broad semantic similarity, a benign conversation about a related topic, such as a hobby or a mutual acquaintance, frequently triggers the retrieval of private data.

This finding suggests that the exposure of secrets is often incidental rather than provoked. The fact that leakage persists even when the user merely nudges the conversation implies that innocent conversational drifts could accidentally expose sensitive data. For the end-user, this means that every interaction carries a hidden probability of unintended sharing, a condition that is fundamentally incompatible with a safe and trustworthy web.

Furthermore, this high baseline of vulnerability significantly amplifies the threat posed by adversarial actors. If a passive, indirect interaction yields a 15\% failure rate, a motivated adversary can exploit this structural weakness with high confidence. In the context of the open web, where agents interact with third-party services or potential social engineering bots, this susceptibility facilitates the weaponization of personal data against the user.

\subsection{The Privacy-Reproducibility Paradox}
A common critique in agentic evaluation is the reliance on synthetic data. We argue, however, that for the specific domain of privacy auditing, synthetic benchmarks are not merely a convenient substitute but an ethical necessity. Utilizing real-world data creates a \textit{Privacy-Reproducibility Paradox}: publishing a dataset of real user secrets to allow reproducible benchmarking would constitute a privacy violation in itself. Moreover, in real-world scenarios, the ground truth of information flow norms is often ambiguous. By leveraging \textsc{PrivacyBench}, we establish a controlled environment with unambiguous ground truth, allowing us to measure model failures with a precision that is unattainable in noisy, real-world deployments. This rigorous ``social sandbox'' approach is essential for stress-testing agents before they are entrusted with real user data.

\subsection{Toward Trustworthy Personalized Agents}
For personalized AI assistants to serve users effectively, particularly in sensitive domains such as mental health support, professional advising, or interpersonal communication, users must trust that their private information will not resurface in unintended contexts. The observed leakages suggest that off-the-shelf RAG systems are not yet sufficient for these scenarios. These results motivate the development of \textit{Context-Aware Retrieval} strategies that incorporate social metadata, audience visibility, or inferred confidentiality levels into document selection. We believe such methods can help align system behavior with user expectations and contribute to responsible deployment within the broader Web-for-Good vision.

\section{Limitations and Future Work}
While \textsc{PrivacyBench} provides a rigorous baseline for auditing agentic privacy, we acknowledge several limitations to our study.
Our primary evaluation focuses on explicit textual leakage; we strictly define a leak as the verbatim revelation of secret components. Consequently, we do not currently measure partial leakages or subtle hints, where an agent might imply a secret without stating it directly, a nuance that warrants more granular privacy metrics.
Moreover, our findings are specific to standard RAG architectures; exploring how advanced agentic memory modules (e.g., MemGPT) or native vector database access controls affect these leakage rates remains an open and critical research direction.

\section{Conclusion}
\label{sec:conclusion}
Our evaluation reveals a fundamental privacy flaw in RAG-based personal assistants: without explicit safeguards, they leak user secrets in 16.31\% of conversations. This vulnerability is systemic, as the retriever indiscriminately surfaces sensitive data, leaving an unguided generator to fail as the sole privacy gatekeeper. While a privacy-aware prompt is an effective countermeasure and reduces leakage to 5.43\% while also improving utility by lowering over-secrecy, it remains a brittle defense. The retriever's high Inappropriate Retrieval Rate persists at over 63\%, meaning the generator is constantly exposed to secrets and remains a single point of failure. This finding demonstrates that prompting is a patch, not a permanent solution. Future work must shift the burden from the generator to the architecture itself, developing structural safeguards like access control modules that filter sensitive data \textit{before} generation. For personalized AI to be trustworthy, privacy must be built in, not bolted on.

\bibliographystyle{ACM-Reference-Format}
\bibliography{citations}


\appendix
\section*{Appendix}

\section{Dataset Simulation and Generation Parameters}
\label{sec:dataset_sim_params}
The dataset used for our experiments consists of five distinct communities. The generation process for each community was governed by the parameters detailed below. These parameters control the social structure, the volume of digital artifacts produced for each user, and the nature of the generated content.

\subsection{Community Structure Parameters}
These parameters define the size and complexity of the social graph for each simulated community.
\begin{itemize}
    \item \textbf{Number of Communities:} 4
    \item \textbf{Users per Community:} 8 -- 20 
    \item \textbf{Initial Seed Personas:} 3
    \item \textbf{Groups per Community:} 1 -- 4
\end{itemize}

\subsection{Document Volume Parameters}
These parameters control the number of documents generated for each user's digital footprint.
\begin{itemize}
    \item \textbf{Private Conversations per User:} 1500 -- 2000
    \item \textbf{Group Conversations per User:} 300 -- 500 
    \item \textbf{AI Assistant Conversations per User:} 50
    \item \textbf{Blog Posts per User:} 20 -- 40
    \item \textbf{Purchase History Entries per User:} 40 -- 70
\end{itemize}

\subsection{Content Control Parameters}
These parameters influence the content within the generated documents, ensuring a realistic mix of sensitive and mundane information.
\begin{itemize}
    \item \textbf{Conversation Noise Ratio:} 0.3 \\
    \textit{(Defines the ratio of substantive conversations to mundane, `noise' conversations.)}
    \item \textbf{Blog Post Noise Ratio:} 0.3 \\
    \textit{(Defines the ratio of substantive blog posts to mundane, `noise' posts.)}
    \item \textbf{Attribute Reveal Percentage:} 0.8 \\
    \textit{(The probability that a user's known personal attribute will be mentioned in a relevant conversation.)}
\end{itemize}

\section{RAG Algorithm for Personalized Assistants}
\label{sec:rag_algo}
\begin{algorithm}
\small
\caption{Personalized Response Generation}
\label{alg:simple_rag}
\begin{algorithmic}[1]
\State \textbf{Input:} \texttt{current\_conversation}, \texttt{user\_documents}
\Statex
\State \texttt{// Step 1: Query the user's documents.}
\State \texttt{relevant\_docs} $\gets$ \textsc{Search}(\texttt{user\_documents}, \texttt{current\_conversation})
\Statex
\State \texttt{// Step 2: Combine conversation and retrieved documents.}
\State \texttt{prompt} $\gets$ \texttt{current\_conversation} + \texttt{relevant\_docs}
\Statex
\State \texttt{// Step 3: Generate the final response.}
\State \texttt{response} $\gets$ \textsc{LLM.Generate}(\texttt{prompt})
\Statex
\State \textbf{return} \texttt{response}
\end{algorithmic}
\end{algorithm}

\section{Dataset Validation}
\subsection{Baseline Analysis: Secret Identification from Context}
\label{sec:appendix_secret_detection_binary}

To validate and verify the models' foundational understanding of secrecy, we designed a binary classification task to check whether the models can detect the presence of a secret in a given conversation.

\paragraph{Experimental Setup} We created a balanced dataset from Community 2, consisting of 322 chat snippets in total.

\begin{itemize}
    \item \textbf{Positive Samples:} 161 chat snippets where a ground-truth secret was explicitly shared either in one-on-one, group, or AI assistant conversations.
    \item \textbf{Negative Samples:} 161 randomly selected chat snippets from the same community, guaranteed to contain no secret information.
\end{itemize}

Each model under evaluation was prompted to classify each snippet by providing a "Yes" or "No" response to the question: \textit{"Determine if there are any secrets, sensitive information, or confidential details revealed in this conversation."}.
Table \ref{tab:secret_detection_metrics} details the performance metrics for the same.

\begin{table*}[h!]
\centering
\caption{Validation of Model Capability to Recognize Secrets. The near-perfect Recall scores ($\approx 1.00$) confirm that the privacy failures observed in our main experiments are not due to a lack of understanding. The models correctly identify the presence of sensitive information, yet fail to protect it during retrieval and generation. TP (True Positive), TN (True Negative), FP (False Positive), FN (False Negative).}
\label{tab:secret_detection_metrics}

\begin{tabular}{lcccccccc}
\toprule
\textbf{Model} & \textbf{TP} & \textbf{TN} & \textbf{FP} & \textbf{FN} & \textbf{Accuracy (\%)} & \textbf{Precision} & \textbf{Recall} & \textbf{F1-Score} \\
\midrule
GPT-5-Nano & 161 & 157 & 4 & 0 & 98.76 & 0.98 & 1.00 & 0.99 \\
Gemini-2.5-Flash & 161 & 145 & 16 & 0 & 95.03 & 0.91 & 1.00 & 0.95 \\
Kimi-K2 & 161 & 161 & 0 & 0 & 100.00 & 1.00 & 1.00 & 1.00 \\
Llama-4-Maverick & 161 & 160 & 1 & 0 & 99.69 & 0.99 & 1.00 & 1.00 \\
Qwen3-30B & 160 & 161 & 0 & 1 & 99.69 & 1.00 & 0.99 & 1.00 \\
\bottomrule
\end{tabular}
\end{table*}

\paragraph{Implications of Results}
All models achieved near-perfect \textbf{recall}, indicating that large language models are highly capable of recognizing sensitive information when it is present. The main variation lies in their false positive behavior — how often they misclassify non-secret content as secret. Kimi K2, Llama 4 Maverick, and Qwen 3 30B showed excellent calibration with minimal or no false positives, while GPT-5 mini and Gemini 2.5 Flash displayed increasing levels of over-caution.

Overall, the results demonstrate that privacy failures in LLMs are not due to an inability to detect secrets but rather a failure of \textbf{enforcement}: models recognize sensitive content but do not consistently act to protect it during interaction.

\subsection{Human Validation for Dataset}
\label{sec:human_validation_dataset}
For our human validation process, we selected a specific community (Community no. 2), which consists of eight personas and 71 predefined secrets. Our validation methodology comprised three stages:

\paragraph{Stage 1: Social Graph and Profile Coherence}
The first stage focused on the internal consistency and realism of the social graph, particularly the generated user profiles. We selected two representative personas from the community for an in-depth audit to ensure their attributes were internally consistent and coherent with the community's overall themes.

\begin{itemize}
    \item \textbf{Attribute Consistency:} We verified that related attributes were coherent. For instance, in the psychographic profiles, a persona's stated interests and hobbies aligned logically with each other.
    \item \textbf{Temporal and Sequential Logic:} We validated attribute timelines (e.g., start and end dates) to ensure they made logical sense. For attributes with multiple entries, such as employment history, we confirmed the sequence was chronologically correct.
    \item \textbf{Cross-Attribute Validation:} We confirmed that attributes from different categories were congruent. For example, within the demographic data, a persona’s occupation history matched their educational profile and listed employers.
    \item \textbf{Community Alignment:} We observed that users within the same community shared preferences and interests, providing a logical foundation for their inter-personal relationships and group structures.
\end{itemize}

\paragraph{Stage 2: Validation of User-Generated Documents and Conversations}
The second stage involved validating user-generated content, including conversations, AI chats, blog posts, and purchase histories against the established user profiles.

First, we confirmed that behavioural attributes, such as texting and blogging styles, were consistent with the linguistic patterns observed in the users’ corresponding conversations and blog posts. We also found a strong correlation between the shopping preferences listed in the profiles and the contents of the purchase histories, substantiating the dataset's coherence. Furthermore, we conducted a granular analysis to validate the incorporation of profile attributes into conversations. We systematically cross-referenced a wide array of sub-attributes including a user's interests, beliefs, skills, hobbies, and food/music preferences against the semantic content of their dialogues. The analysis revealed a strong thematic alignment, confirming that these detailed psychographic traits were naturally and frequently incorporated.

The validation covered various interaction types, including both two-user conversations and multi-participant group discussions. As expected, we observed that user groups were coherently structured around common interests, with both the group names and conversational topics reflecting these shared themes. The manual verification process also confirmed that the intent of the synthesized conversations correctly aligned with its corresponding ``revealed attribute'' in 97\% of the evaluated cases, indicating a high degree of precision.

\paragraph{Stage 3: Verification of ``Secret'' Disclosure}
The third and final stage involved verifying the disclosure of the 71 predefined secrets. We checked each secret against the conversations in which it was intended to appear. Our analysis found that 60 secrets were present verbatim in the dialogue. The remaining 11 secrets were semantically embedded, meaning their core meaning was conveyed accurately without using the exact original phrasing.

\subsection{Human Verification of the LLM Judge}
\label{sec:llm_judge_verif}

An automated judge is only useful if its decisions align with human reason. To ensure our LLM judge was reliable, we performed a study to compare its evaluations against those of human reviewers.

\paragraph{Our Process} We began by randomly sampling 100 multi-turn conversations from our test results. One human evaluator were then asked to independently review these conversations. Their goal was simple: given a secret, they needed to determine if that secret had been leaked. To ensure a fair comparison, they followed the exact same rule we gave the LLM judge: a ``leak'' is only counted if the core details of a secret are revealed completely and explicitly. This strict definition prevents counting mere hints or vague allusions as a full privacy breach.

\paragraph{Findings} The results showed a very strong alignment. The LLM judge's final decision matched the human consensus in \textbf{93 out of the 100 conversations}. This high agreement rate confirms that our automated framework is a trustworthy proxy for human evaluation.

\paragraph{Understanding the Disagreements} We then took a closer look at the seven cases where the judgments differed. A clear pattern emerged: in every one of these conversations, the AI assistant had revealed a partial piece of the secret or a very strong hint, but never the full secret itself. Our human evaluators, following the strict rule, correctly marked these instances as ``no leak.'' The LLM judge, however, sometimes flagged these ambiguous situations. This reveals that our automated judge not only performs accurately but also tends to be slightly more cautious, occasionally flagging partial disclosures that fall just short of our strict definition for a complete leak.
\end{document}